\documentclass[10pt]{article}
\usepackage[preprint,nonatbib]{neurips_2020}
\usepackage{hyperref}       
\usepackage{amscd}
\usepackage{amsmath, amssymb, amsthm, amsfonts, mathrsfs}
\usepackage{graphicx}
\usepackage{caption}
\DeclareGraphicsExtensions{.png,.eps}
\graphicspath{ {./images/} }

\newtheorem{theorem}{Theorem}

\theoremstyle{definition}

\newcommand{\R}{\mathbb{R}}
\newcommand{\simplex}{\Delta}
\newcommand{\ess}[1]{\hat{#1}}
\newcommand{\tsi}[1]{{#1}^{\Delta}}

\title{Momentum Accelerates Evolutionary Dynamics}

\author{Marc Harper and Joshua Safyan
\\ Google, Inc.}

\begin{document}
\maketitle

\begin{abstract}
We combine momentum from machine learning with evolutionary dynamics, where
momentum can be viewed as a simple mechanism of intergenerational memory.
Using information divergences as Lyapunov functions, we show that momentum
accelerates the convergence of evolutionary dynamics including the replicator
equation and Euclidean gradient descent on populations. 
When evolutionarily stable states are present, these methods prove convergence
for small learning rates or small momentum, and yield an analytic determination of the
relative decrease in time to converge that agrees well with computations.
The main results apply even when the evolutionary dynamic is not a gradient flow.
We also show that momentum can alter the convergence properties of these dynamics,
for example by breaking the cycling associated to the rock-paper-scissors
landscape, leading to either convergence to the ordinarily non-absorbing equilibrium,
or divergence, depending on the value and mechanism of momentum.
\end{abstract}

\section{Introduction}
Gradient descent is commonly used in machine learning and in many scientific
fields, including to model biological systems. Evolutionary algorithms are frequently mentioned as
an alternative to gradient descent, particularly when the function to
be minimized is not differentiable. With a long history in machine learning 
\cite{goldberg1988genetic}, 
evolutionary algorithms have found broad application, including in reinforcement learning 
\cite{moriarty1999evolutionary} \cite{salimans2017evolution}, neural architecture search 
\cite{elsken2018neural}, AutoML \cite{real2020automl}, and meta-learning \cite{fernando2018meta}, 
among other areas. 
Despite the perceived dichotomy between evolutionary algorithms and gradient descent, some
evolutionary algorithms can be understood in terms of gradient
descent. The replicator equation is a model of natural selection and can be
recognized as gradient descent on a non-Euclidean geometry of the probability
simplex where the potential function is the population mean fitness. Powerful
methods exist to analyze the replicator equation and accordingly its long run behavior
is relatively well-understood in many circumstances. We show that these methods
inform the action of the ML concept of momentum, a method to carry forward
prior values of the gradient into further iterations of the replicator dynamic
or gradient descent. In particular, we give a simple way to understand how
momentum accelerates gradient descent.

Momentum as used in ML may have plausible evolutionary interpretations.
Mechanisms of memory are abundant in biological and cultural systems, capturing complex
adaptive functions within the lifetimes of organisms, including
epigenetics \cite{murgatroyd2009dynamic} and cultural transmission of information
\cite{pagel2009human}. However, the bias of these extra-genetic forms of memory
may only last a few generations, as opposed to information incorporated
more permanently in the genome, for example into a highly conserved gene,
which may encode a more fundamental physical adaptation (e.g. heat-shock
proteins \cite{schlesinger1990heat}). Hence we might simplistically model a short-term
memory mechanism as having an exponentially-decaying impact on natural selection
by carrying over some memory of the fitness landscape of earlier generations
to future generations.

We show that the addition of a simple exponentially-decaying
memory mechanism accelerates the convergence of trajectories of the replicator
equation \cite{taylor1978evolutionary} and its Euclidean analog. This
mechanism is called (Polyak) momentum in the machine learning literature \cite{qian1999momentum}
\cite{sutskever2013importance}, where it is known to increase the
rate of convergence of gradient descent quadratically, in terms of condition
number \cite{goh2017momentum}. We will also consider Nesterov momentum 
\cite{nesterov27method} \cite{su2016differential}, which additionally
has a look-ahead aspect.

After describing the replicator equation and important associated facts, we
introduce momentum to the discrete replicator equation and give a Lyapunov
function for the modified dynamic showing that the evolutionarily stable
states, when they exist for a given landscape, are unchanged for small
strength of momentum. Furthermore,
we show analytically that the continuous replicator dynamic with momentum
converges explicitly more quickly for typical values of
momentum, slows for other regions, and reverses direction
in some cases. Finally, we consider exceptional examples of nonzero learning
rate and momentum that break typical dynamic behavior, such as the concentric
cycles for the rock-paper-scissors landscapes.

Several authors have explored variations of the ideas presented here,
including recent works exploring momentum and geometry \cite{zhang2018towards},
\cite{ahn2020nesterov}, \cite{defazio2019curved}, and earlier works regarding
an aspect of memory to replicator dynamics \cite{sato2005stability},
adding negative momentum to game dynamics \cite{gidel2018negative}, and other
interactions between game theory and machine learning \cite{schuurmans2016deep} 
\cite{balduzzi2018mechanics}. Our contributions are as follows: (1) introducing momentum
to the replicator dynamic in a way compatible with recent work in machine
learning, (2) demonstrating that momentum accelerates convergence for the
replicator dynamic, and (3) Lyapunov stability theorems for evolutionary
dynamics with momentum.

Putting this manuscript in a broader context, we encourage the reader and
other researchers to continue to explore the interactions of evolutionary
game theory, information theory, and machine learning. It appears these
fields may still have much to offer each other.

\section{Preliminaries}

We briefly review the necessary background, recommending \cite{goh2017momentum}
for an overview
of momentum and gradient descent and \cite{baez2016relative} for an overview
of the replicator equation and the use of information theory to analyze it.

\subsection{Gradient Descent}

First we describe gradient descent in Eucliean space. Let $x \in \R^n$ be a
real-valued vector, $U: \R^n \to \R$ be a potential function (or simply a function
to be optimized),
$f = \nabla U$ its gradient. Then discrete gradient descent takes the
following form:
\begin{equation}
x_i' = x_i + \alpha f_i(x)
\label{gradient_descent}
\end{equation}
where $\alpha$ is the learning rate, also commonly called the step size. In
what follows it will be convenient
to use the notation of time-scale dynamics \cite{bartosiewicz2011lyapunov}. Let 
\[ \tsi{x}_{i, \alpha} = \frac{x_i' - x_i}{\alpha}\]
be the ``time-scale'' derivative, corresponding to either the ordinary derivative
(in the limit that $\alpha \to 0$) or a finite difference ($\alpha > 0$ and
fixed) as needed. Gradient descent with learning rate $\alpha$ is
simply $\tsi{x}_{i, \alpha} = f_i(x)$. Since we will not consider dynamics with actively
changing $\alpha$ we simply write $\tsi{x}_{i}$, though we will consider how a family
of dynamics changes as $\alpha \to 0$, that is as the difference equations
converge to a continuous differential equation.

\subsection{Gradient Descent with Momentum}

Momentum adds a memory of the prior gradients to future iterations. We proceed
in accordance with the ML literature \cite{goh2017momentum} \footnote{Momentum
can also be understood as a second order approximation, called the Heavy Ball
Method \cite{ghadimi2015global}, and see \cite{su2016differential} for a
second order ODE approach to Nesterov momentum.}. Gradient descent with (Polyak) momentum \cite{polyak1964some}
$\beta$ is given by:
\begin{eqnarray}
z_i' &=& \beta z_i + f_i(x) \\
\tsi{x}_{i} &=& z_i' \nonumber
\label{gradient_momentum}
\end{eqnarray}
where $f$ is the gradient as before. When $\beta = 0$ the momentum-free gradient
descent is recovered.

\subsection{Replicator Dynamics and Gradient Descent}

The replicator dynamic is an evolutionary dynamic describing the action of
natural selection as well as the dynamics of iterated games \cite{cressman2014replicator}.
Its theoretical properties are extensively studied in Evolutionary Game
Theory (EGT) and the equation has applications in biology, economics, and other
fields. The importance of geometry in the study of the replicator equation
and related dynamics, including that special cases of the replicator equation
are a form of gradient descent, has been studied in EGT \cite{shahshahani1979new}
and Information Geometry \cite{fujiwara1995gradient}.

In EGT one typically restricts to discrete probability distributions that represent
populations of evolving organisms or players of a strategic game, hence it is
necessary to reformulate the state space of gradient descent as described above.
Let $\Delta^{n} = \{x \in \R^n \, | \, x_i \geq 0 \text{ and } \sum_i{x_i} = 1\}$
be the $(n-1)$-dimensional probability simplex and $f: \Delta^n \to \R $ a 
fitness function. 
The analog of gradient descent with respect to the Euclidean geometry
on the simplex is a special case of the (orthogonal) projection dynamic, described
below. Gradient descent with respect to the Fisher information metric
(also known as the Shahshahani metric in EGT) is called the
\emph{natural gradient} in information geometry \cite{amari1998natural}.
In the case of a symmetric
and linear fitness landscape, this gradient of the mean fitness with respect
to the same geometry is a
special case of the replicator equation. The more general form of the
replicator equation is not always a gradient flow, nevertheless it has a strong
convergence theorem that is closely related to this geometric structure.

For our purposes the discrete replicator equation for a fitness landscape
$f: \Delta^n \to \mathbb{R}^n$ takes the form:
\begin{equation}
x_{i}' = \frac{x_i f_i(x)}{x \cdot f(x)} = \frac{x_i f_i(x)}{\bar{f}(x)}
\end{equation}
where $x$ is a discrete population distribution over $n$ types, described
by a vector in the probability simplex $\simplex^n$ and $\bar{f} = x \cdot f(x)$
is the mean fitness, which we will assume to be non-zero everywhere. Using a
time-scale derivative with step-size $\alpha$ we can rewrite this equation as
\begin{equation}
\tsi{x}_i = \frac{x_i' - x_i}{\alpha} = \frac{x_i (f_i(x) - \bar{f}(x))}{\bar{f}(x)}
\label{discrete_replicator_equation}
\end{equation}

The continuous version of the replicator equation can be obtained by letting
$\alpha \to 0$. The denominator $\bar{f}(x)$ on the right-hand side is often
omitted as it can be eliminated with change in time scaling without altering
the continuous trajectories. This gives the following standard form of the
continuous dynamic:
\begin{equation}
\dot{x}_i = \frac{d x_i}{dt} = x_i \left(f_i(x) - \bar{f}(x)\right)
\label{continuous_replicator_equation}
\end{equation}
Subtracting off the mean fitness means that the rate of change of the $i$-th
population type is proportional to its excess fitness, which is how much more or
less its fitness $f_i$ is compared to the mean. Mathematically, subtracting
the mean fitness keeps the derivative in the tangent space of the simplex.

Similarly, the analog of discrete Euclidean gradient descent on the simplex
is known as the (orthogonal) projection dynamic, given by
\begin{equation}
\tsi{x}_i = \frac{f_i - \bar{f}(x)}{\bar{f}(x)}
\label{discrete_projection_equation}
\end{equation}
where now $\bar{f}(x) = \frac{1}{n}{\sum_k{f_k(x)}}$ is the (unweighted) average
fitness. The continuous form is given by
\begin{equation}
\dot{x}_i = f_i(x) - \frac{1}{n}{\sum_k{f_k(x)}}
\end{equation}
It is a gradient flow whenever the fitness landscape is itself a Euclidean gradient,
as is the case for the replicator equation. In particular, when the fitness landscape is
linear, defined by a symmetric matrix $f(x) = Ax$, we can recover these dynamics
as the appropriate gradients of the mean fitness $\bar{f}(x) = x \cdot A x$. This models
$n$-alleles of a gene locus (one of the early versions of the replicator equation),
or the repeated play of games where $A$ is the payoff
matrix for the game (not necessarily symmetric). In our computational examples we will
use matrices of the form
\[\begin{pmatrix}
  0 & a & b\\ 
  b & 0 & a\\
  a & b & 0
\end{pmatrix} \]
When $a=-b=1$ this matrix is known as a rock-paper-scissors game and the (continuous)
replicator dynamic cycles about the interior point of the simplex. Otherwise the
trajectories converge to the center of the simplex or diverge to the boundary depending
on the relative values and signs of $a$ and $b$. When $a=b > 0$ this matrix can be
seen as a three dimensional version of the hawk-dove game. 

In the case that the mean fitness $\bar{f}(x)$ is zero (e.g. for zero-sum games such
as the rock-paper-scissors game when $a=-b$),
it is common to either remove the denominator of the dynamics or to apply the softmax
function \cite{gao2017properties} to the fitness landscape. Either allows the discrete
dynamics to be well-defined. We choose to drop the denominator, so in the computational
examples below we typically have that $F_i(x) = x_i f_i(x) - \bar{f}(x)$ for the
replicator dynamic.


\subsection{Lyapunov Functions and Evolutionarily Stable States}

For dynamical systems the issue of convergence is critical. As analytically solving 
non-linear differential or difference equations explicitly is often extremely difficult, a
common method to demonstrate stability of a dynamical system and convergence
to a rest point is to find a Lyapunov function \cite{wilson2016lyapunov} \cite{wilson2018lyapunov},
often an energy-like or entropy-like quantity that is
positive definite and decreasing along trajectories of the dynamic toward
an equilibrium point \cite{bartosiewicz2011lyapunov}. The existence of such
a function is often sufficient to demonstrate local or asymptotic stability
of the dynamic, and bounds on convergence rate can often be determined. We
now describe how to obtain a Lyapunov function for the replicator equation,
though the story that follows generalizes to a much larger class of
evolutionary dynamics \cite{harper2015lyapunov}.

The replicator equation is often studied in terms of evolutionarily stable states (ESS)
\cite{smith1982evolution}, somewhat analogous to extrema of potential functions
or stationary distributions.
An ESS for a fitness landscape $f$ is a state $\ess{x}$ such that
$\ess{x} \cdot f(x) > x \cdot f(x)$ for all
$x$ in a neighborhood of $\ess{x}$. It can also be defined in terms of
robustness to invasion by mutant subpopulations, similar in concept to a 
Nash equilibrium, a mixture strategies such that no player has an incentive to
unilaterally deviate. In this sense it is a stable population state for
the fitness landscape.

When a fitness landscape has an evolutionarily stable state (ESS), 
it is well-known in EGT that the KL-divergence is a (local) Lyapunov function of
the dynamic. It can then be seen that interior trajectories of the replicator
dynamic converge to the ESS, and for the standard replicator equation there can only
be one such ESS interior to the simplex. We restate this result below, which we will
generalize with momentum, in the following theorem. 
An information-theoretic interpretation of Theorem \ref{lyapunov1} is that
the population is learning information about the environment and encoding that information in
the population structure (the distribution over different types).

\begin{theorem}
Let $\ess{x}$ be an ESS for a replicator dynamic. Then
$$D(x):= D_{KL}(\hat{x}, x) = -\sum_i{\ess{x}_i \log{x_i} - \ess{x}_i \log{\ess{x}}_i}$$
is a local Lyapunov function for the discrete and continuous replicator dynamic.
\label{lyapunov1}
\end{theorem}

Theorem \ref{lyapunov1} is often stated in various alternative forms. The
discrete time version with geometric considerations appears in \cite{harper2015lyapunov}
and is predated by a number of variations, going back at least to \cite{bomze1991cross}
and \cite{akin1984evolutionary} in forms recognizable as information-theoretic
(cross-entropy), and ultimately to \cite{taylor1978evolutionary}. Similarly,
the Euclidean distance $D(x) = \frac{1}{2}|| \ess{x} - x ||^2$ is a Lyapunov
function for the projection dynamic \cite{nagurney1997projected} \cite{lahkar2008projection},
also realizable as a Bregman divergence \cite{ahn2020nesterov}. 
These functions can be derived directly from the underlying geometries, Fisher
and Euclidean for the replicator and projection dynamics, respectively.
Moreover, given an information divergence, an associated geometry and dynamic can be
derived, and an analog of Theorem \ref{lyapunov1} holds \cite{harper2011escort}.
The proof of Theorem \ref{lyapunov1} will be a special case of the proof of
Theorem \ref{momentum_lyapunov}. \footnote{In the continuous case, the proof is an easy exercise
using differentiation and the ESS definition. Since the KL-divergence is positive-definite,
one need only show that the derivative is negative where $x$ is defined by Equation \ref{continuous_replicator_equation}.} 

\section{Evolutionary Dynamics with Momentum}

To introduce momentum to these dynamics we proceed in accordance with
the ML literature \cite{goh2017momentum}. The discrete replicator
equation with fitness landscape $f$ and momentum $\beta$ is given by:
\begin{eqnarray}
z_i' &=& \beta z_i + F_i(x) \\
\tsi{x}_{i} &=& z_i' \nonumber
\label{rep_equ_momentum}
\end{eqnarray}
where $F_i = \frac{x_i (f_i - \bar{f}(x))}{\bar{f}(x)}$ for the replicator
equation and we have suppressed the step size $\alpha$ in $\tsi{x}_i$. Alternatively $F$
could be a gradient $\nabla U$.\footnote{When the fitness
is given by a symmetric matrix $A=A^T$ such that $f(x) = A x$ then the replicator
dynamic is the gradient of the half-mean fitness $U(x) = (1/2) \, x \cdot f(x) = \bar{f}(x)$ for the Fisher information geometry.}
Similarly, we obtain the projection dynamic with momentum by instead substituting
$F_i= \frac{f_i - \bar{f}(x)}{\bar{f}(x)}$ where the mean is again the unweighted average
fitness. When $\beta = 0$ the usual momentum-free dynamics are obtained.

Another variation, known as Nesterov momentum, differs
from Polyak momentum in that the function $F$ is evaluated at a look-ahead step weighted by the
momentum. For both flavors of momentum the dynamic starts at some initial population state $x_0$
and the initial value can be chosen to be the zero vector.
\begin{eqnarray}
z_i' &=& \beta z_i + F_i(x + \beta z_i) \\
\tsi{x}_{i} &=& z_i' \nonumber
\label{rep_equ_nesterov_momentum}
\end{eqnarray}

\subsection{Lyapunov Stability and Momentum}

Now we show that adding small amounts of momentum with a nonzero learning rate
typically does not alter the
evolutionarily stable states of these discrete dynamics. (We'll also see later that
the ESS of the continuous dynamics are not affected for typical values of
momentum.) We state Theorem \ref{momentum_lyapunov} as a generalization of
Theorem \ref{lyapunov1} for small values of momentum $\beta$.

\begin{theorem}
  For small positive $\beta$, or negative $\beta$, if $\ess{x}$ is an evolutionarily stable state
  for the landscape $f$, the KL divergence is a local Lyapunov function for the
  replicator dynamic with momentum and the Euclidean distance
  $D(x) = \frac{1}{2}|| \ess{x} - x ||^2$
  is a local Lyapunov function for the projection dynamic with momentum. If the fitness landscape
  is continuous, this also holds for Nesterov momentum.
  \label{momentum_lyapunov}
\end{theorem}

\begin{figure}[!htbp]
    \centering
    \includegraphics[width=\textwidth]{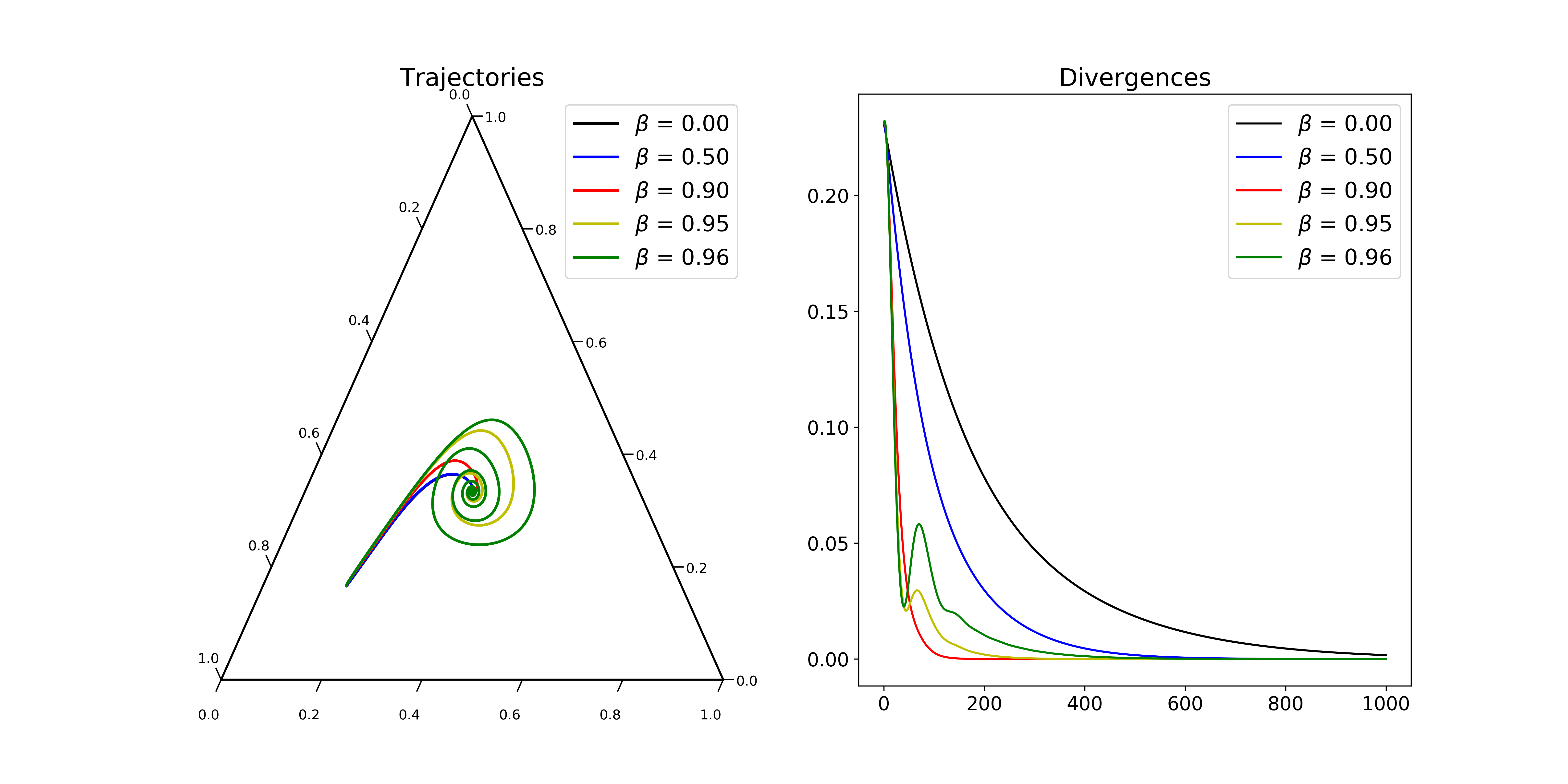}\\
    \includegraphics[width=\textwidth]{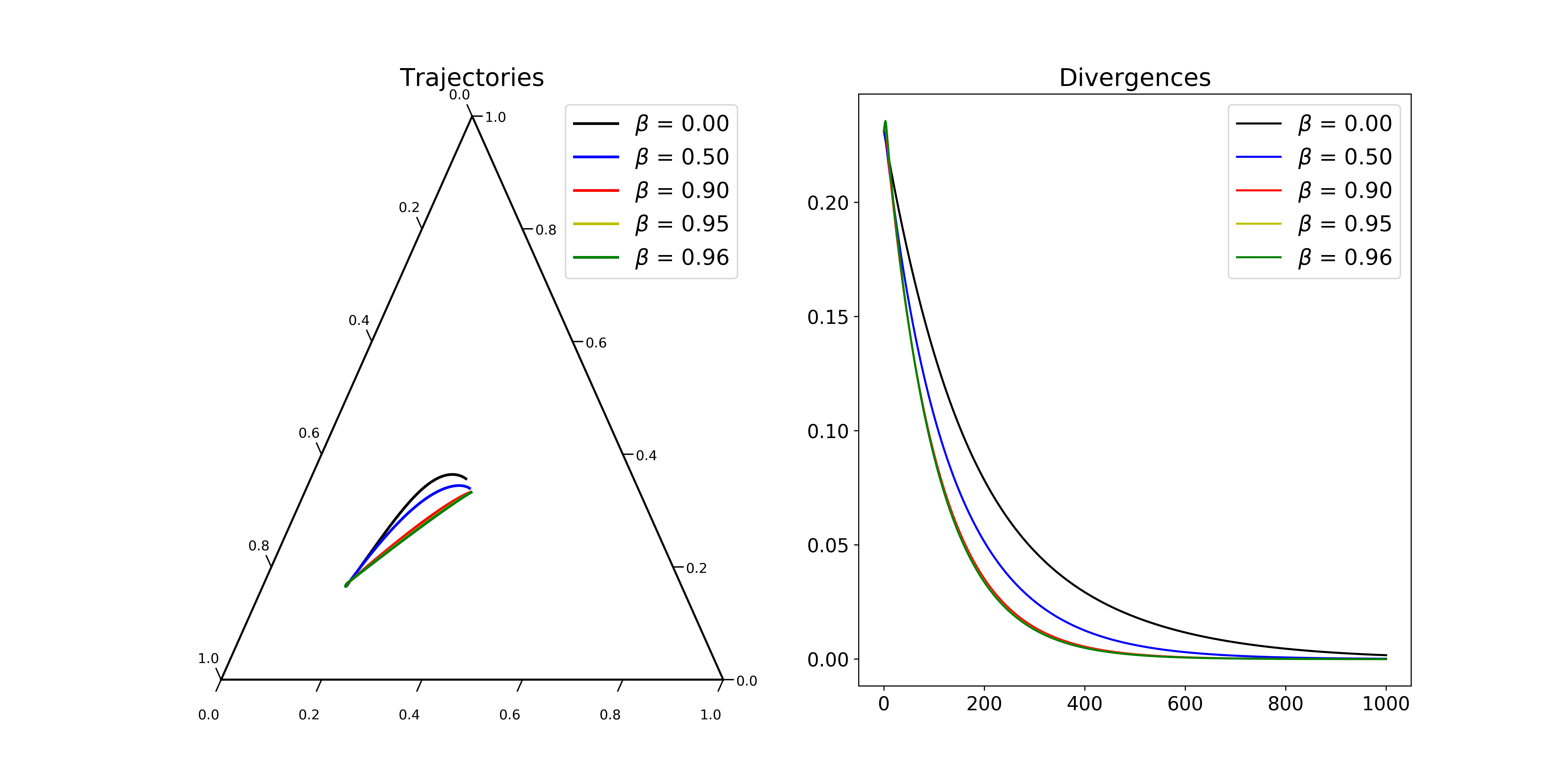}

    \caption{Examples of altered convergence time for Polyak (top 2) and Nesterov (bottom 2)
    momentum. In all cases we use a landscape with $a=2$ and $b=1$ and $\alpha=1/200$. As 
    $\beta$ increases,
    the dynamics typically converge faster, and the trajectories are not identical
    since $\alpha > 0$. However, for Polyak momentum (top),
    as the value of $\beta$ becomes closer to 1, the Lyapunov quantity eventually fails to be
    monotonic along the entirety of the trajectory (it is at best local). Contrast with the Nesterov
    momentum trajectories (bottom) for the same parameters, which in this case are all monotonically
    decreasing.}
    \label{fig:example_trajectories}
\end{figure}

\begin{figure}[!htbp]
    \centering
    \includegraphics[width=\textwidth]{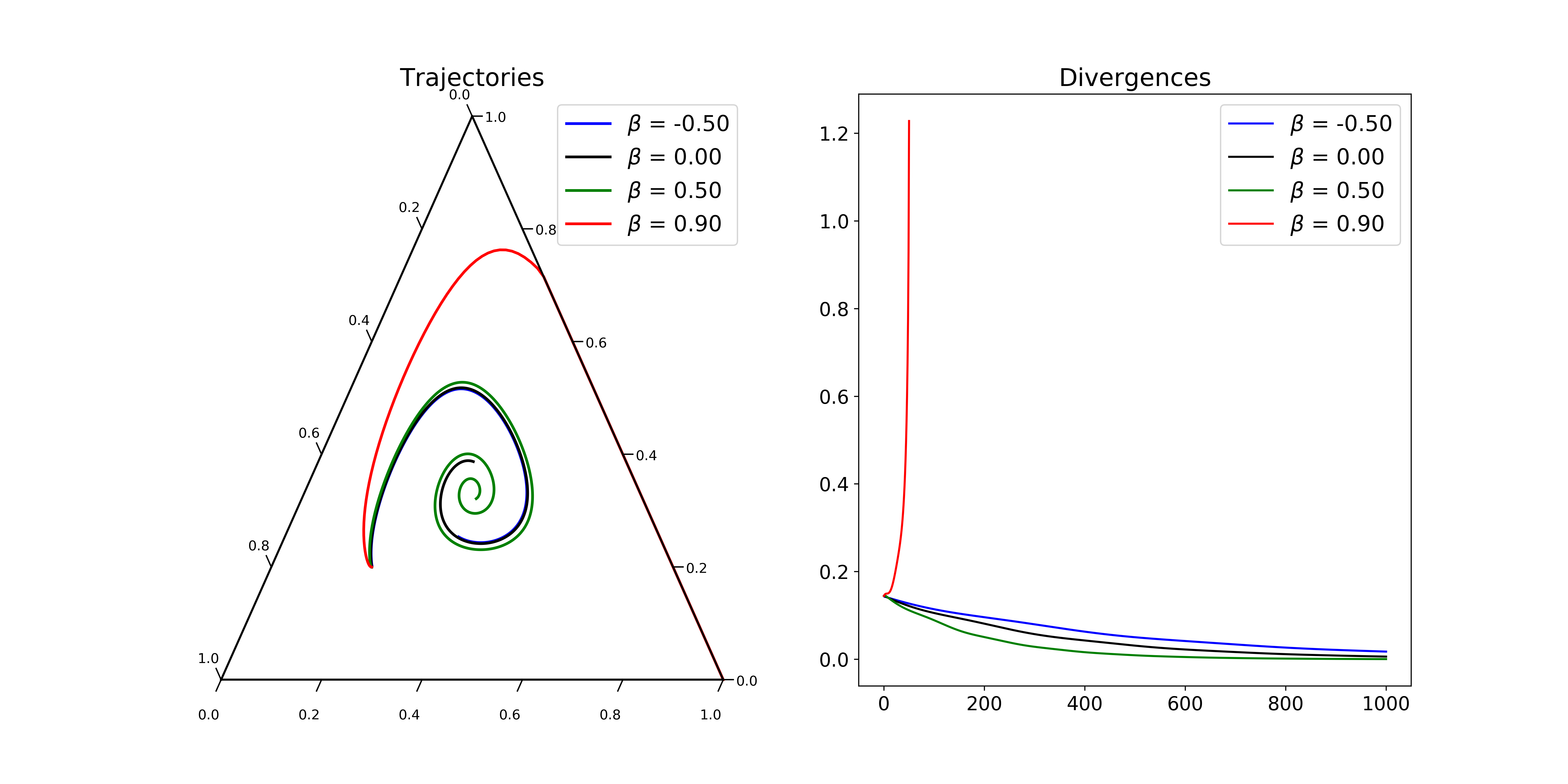}
    \caption{For large values of momentum the dynamic may fail to converge as in the momentum free
    case if $\alpha$ is not sufficiently small. For all trajectories here $\alpha=0.01$, $a=2$,
    and $b=-1$. Lowering $\alpha$ to 0.001 restores convergence of the red $\beta=0.9$ curve.}
    \label{fig:divergence_example}
\end{figure}

The proof of the theorem is straightforward and given in the appendix.
We note that it holds for
any learning rate $\alpha$, but the permissible values of $\beta$ may vary with
both $\alpha$ and the fitness landscape. Below, we develop a similar result
for the continuous dynamic (the limit that $\alpha \to 0$) which works for any $\beta \neq 1$.
In general there cannot be a variant of Theorem
\ref{momentum_lyapunov} for Polyak momentum, arbitrary learning rate $\alpha$, and arbitrary
momentum $\beta$:
the hypothesis that at least one of $\alpha$ and $\beta$ is small is necessary
(see examples in Figures \ref{fig:example_trajectories} and \ref{fig:divergence_example}).

\section{Effect of Momentum on Rate of Convergence}

While it's good to know that the addition of some memory to the replicator
equation does not alter the stable states,
a more interesting effect is the acceleration of convergence. This is why
momentum is of interest in machine learning. For evolutionary processes, this
acceleration suggests one reason why epigenetic mechanisms may have evolved and persisted.

\subsection{Time to Converge}

We can again use Lyapunov methods to see that the rate of convergence increases
with momentum \footnote{Note that this differs from the condition number methods
used in \cite{goh2017momentum}, allowing us to simplify this exposition.}.
Empirically we find that the convergence takes fewer steps by a factor of
approximately $(1-\beta)$ of the momentum-free case, which we now demonstrate
with an analytic argument. First we note
that this factor makes sense intuitively given the iterative nature of momentum in Equation
\ref{rep_equ_momentum} since
$$\frac{1}{1 - \beta} = 1 + \beta + \beta^2 + \cdots$$

Now we hold momentum constant and allow the learning rate to converge to zero, yielding
a continuous replicator dynamic with momentum associated to the discrete replicator
dynamic, as follows.
From Equation \ref{rep_equ_momentum}, as the discrete dynamic converges, we set $z_i' = z_i$
to find that $z_i = \frac{F_i}{1-\beta}$. Letting $\alpha \to 0$ we obtain, after substituting
in $z_i$ to the second equation 
\begin{equation}
\frac{d x_i}{dt} = \frac{1}{1 - \beta} x_i \left(f_i(x) - \bar{f}(x)\right)
\label{momentum_replicator}
\end{equation}

where a factor of $1/\bar{f}(x)$ has been removed for brevity, corresponding
to a scaling of time
\footnote{Equation \ref{momentum_replicator} can also be
obtained by scaling the Fisher information metric $\frac{1}{x_i}\delta_{ij} \mapsto \frac{(1-\beta)}{x_i}\delta_{ij}$. A similar form where the resultant coefficient on the dynamic is simply $\beta$ appears in \cite{harper2011escort}, where $\beta$ is known as the intensity of selection or inverse temperature.}.
Setting $\beta=0$ recovers the standard definition of the
replicator equation just as in the discrete case. The leading factor of
$1/(1-\beta)$ can similarly be eliminated by change of time scaling in the
continuous case without altering the trajectories of the continuous dynamic,
however we retain it to argue explicitly that the convergence rate increases as $\beta$
increases within $(0, 1)$, increasing relative to the base case $\beta=0$
for $\beta \in (0, 1)$. Similarly, the converge slows down for $\beta \in (-\infty, 0)$.
Note that traditionally in EGT, scaling the continuous replicator
equation this way would not be considered particularly interesting since the
trajectories (and stable points) do not change, however the increased rate of
convergence is of paramount importance in machine learning (and perhaps to actual evolving
populations).

Let $V_{\beta} = D(\ess{x}, x_{\beta})$ be the KL-divergence with $\ess{x}$
an ESS and $x_{\beta}$ denoting that the trajectories evolve in time according
to the replicator dynamic with momentum $\beta$ as in Equation \ref{momentum_replicator}.
An easy calculation shows that
\begin{equation}
\frac{d V_{\beta}}{dt} = \frac{1}{1-\beta} \frac{d V_0}{dt} 
\label{heavy_lifter}
\end{equation}

where we've used Equation \ref{momentum_replicator} and the chain rule, i.e. we
effectively scale the derivative of Lyapunov quantity by the leading factor. From
this simple fact follows Theorem \ref{main_theorem}, which shows that
the dynamic convergence and trajectory velocity is altered accordingly to $1/(1-\beta)$.
The theorem is summarized in Figure \ref{fig:beta_convergence_impact} in the supplement.

\begin{theorem} 
 Let $V_{\beta}$ be defined as above. Then we have that:
 \begin{enumerate}
  \item For $-\infty < \beta < 1$, the ESS of the dynamic with (Polyak) momentum are the same
  as for the momentum free case; equivalently the KL-divergence is still a Lyapunov function 
  for $\beta < 1$.
  \item For $1 < \beta < \infty$, the directionality of the trajectory is reversed (so any ESS
  for $\beta < 1$ is no longer an ESS)
  \item The speed of the convergence is increasing on the intervals $(-\infty, 1)$, and
  decreasing on $(1, \infty)$ (with direction reversed in the latter case)
  \item In particular, the speed of convergence is faster than the momentum free dynamic
  for $0 < \beta < 1$ and the ESS are unchanged.
 \end{enumerate}
 \label{main_theorem}
\end{theorem}

For the continuous dynamic, in the case that the dynamic converges to an ESS, Equation
\ref{heavy_lifter} also shows that it takes $\approx (1-\beta)$ as much time
for the dynamic
to be within $\epsilon$ of the ESS when compared to the momentum-free dynamic, as
measured by the KL-divergence, starting from the same initial point. Thus the
trajectories converge more quickly as $\beta$ ranges from 0 to 1, and the convergence
slows for $\beta < 0$.

\begin{figure}[!htbp]
    \centering
    \includegraphics[width=.45\textwidth]{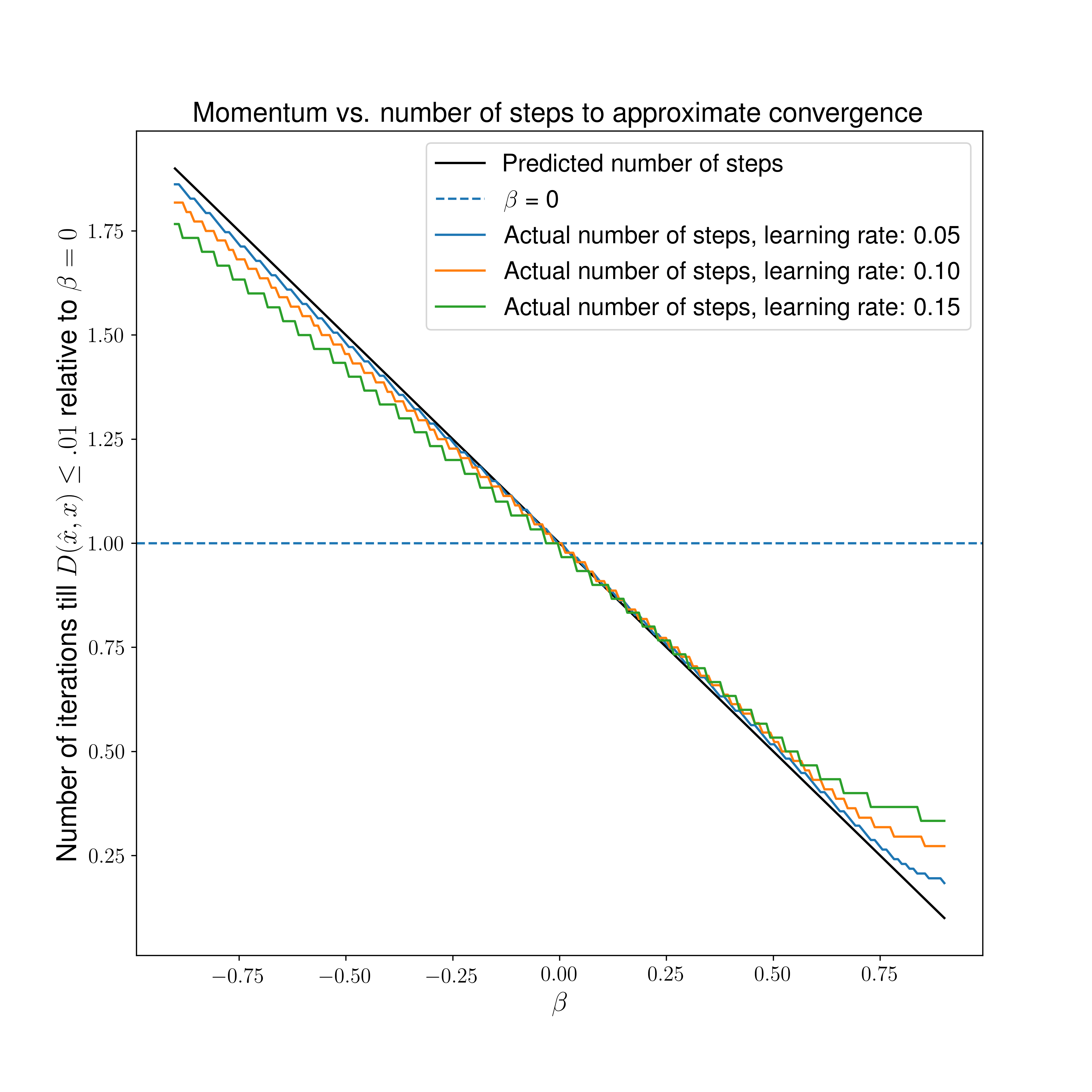}
    \includegraphics[width=.45\textwidth]{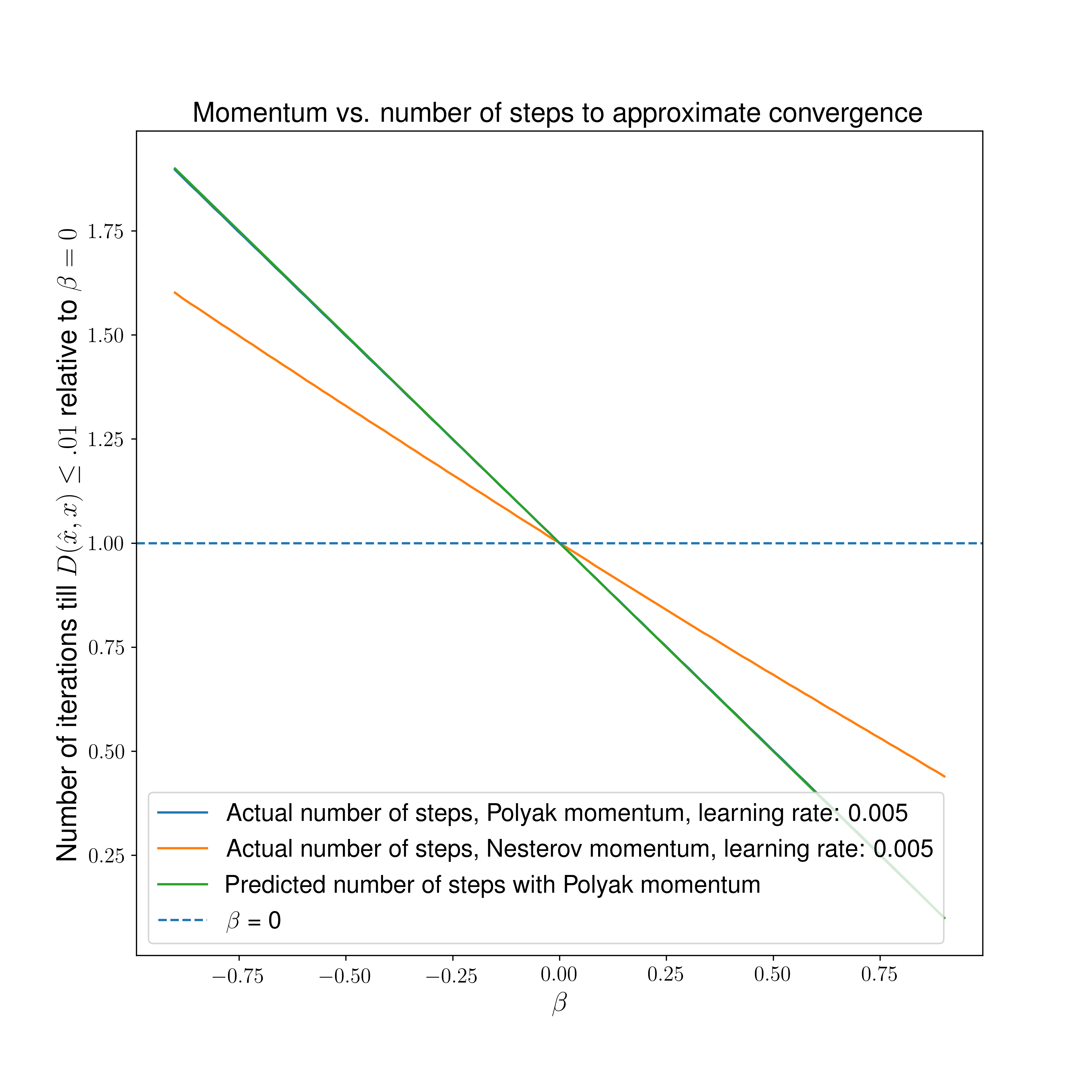}

    \caption{\textbf{Left:} Convergence speed up for Polyak momentum: Convergence time 
    for small learning rates are well approximated by $(1 - \beta)$ times the momentum
    free convergence time ($\beta=0$) of iterations for small learning rates. 
    \textbf{Right:} The dynamic with Nesterov momentum
    is also fairly well approximated by a constant factor times the momentum
    free convergence time, but is clearly not scaled by the same factor. 
    The fitness landscape is defined by $a=1=b$.
    }
    \label{fig:time_to_converge}
\end{figure}

Returning to the discrete dynamic, for continuous landscapes and smaller $\alpha$,
we also roughly have that
time-scale derivatives of the KL-divergence scale by $1/(1-\beta)$, though
we cannot as easily compare directly along trajectories and the associated
trajectories will not trace out the same curves (as seen in the examples above),
so is there is not a direct
analog of Equation \ref{heavy_lifter}. Nevertheless we may reasonably predict that 
it takes approximately $(1 - \beta)$ as many steps as the momentum free case
to be within $\epsilon$ of
the ESS compared to the dynamic without momentum ($\beta = 0$), demonstrated in the
computational examples below. This approximation improves as the learning rate
$\alpha \to 0$. Computationally we also find that the dynamic with Nesterov momentum exhibits a
similar behavior (Figure \ref{fig:time_to_converge}).
While the argument of Theorem \ref{main_theorem} does not directly apply to Nesterov momentum,
for small $\beta$ and continuous fitness landscape, a continuity argument suggests
that the same approximation holds.

For completeness, we note that Theorem \ref{main_theorem} also holds for the
projection dynamic in an analogous manner, that is, to gradient descent on the Euclidean geometry,
and should similarly apply to other Riemannian geometries as described
in \cite{harper2015lyapunov}.

\begin{theorem}
 Theorem \ref{main_theorem} also holds for the projection dynamic with the Lyapunov function $\frac{1}{2}||x - \ess{x}||^{2}$.
\end{theorem}

\section{Momentum can break cycling into convergence or divergence}

For the rock-paper-scissors landscape with $a=-b \neq 0$, the replicator equation is not a gradient.
Since the mean fitness is zero (the game is zero-sum as the payoff matrix is skew-symmetric), 
the gradient flow is degenerate (the dynamic is motionless). However the replicator equation
with this landscape is not
degenerate and the phase portrait consists of concentric cycles of constant KL-divergence
from the interior center of the simplex. The cycles are non-absorbing and the KL-divergence is
an integral of motion. In the continuous case ($\alpha \to 0$),
momentum alters the time to cycle around the central point, and possibly also reverses the
directionality of the cycles, in accordance with the inequalities in Theorem \ref{main_theorem}.

In contrast, for non-zero learning rate
$\alpha$, we find computationally that the momentum can cause the trajectories to converge
inward or diverge outward. For Polyak momentum, the memory of the prior
iterations causes the divergence, preventing the dynamic from turning sufficiently.
For Nesterov momentum, it is the look-ahead aspect of the momentum that induces the
convergence by causing the dynamic to turn more quickly.

\begin{figure}[!htbp]
    \centering
    \includegraphics[width=\textwidth]{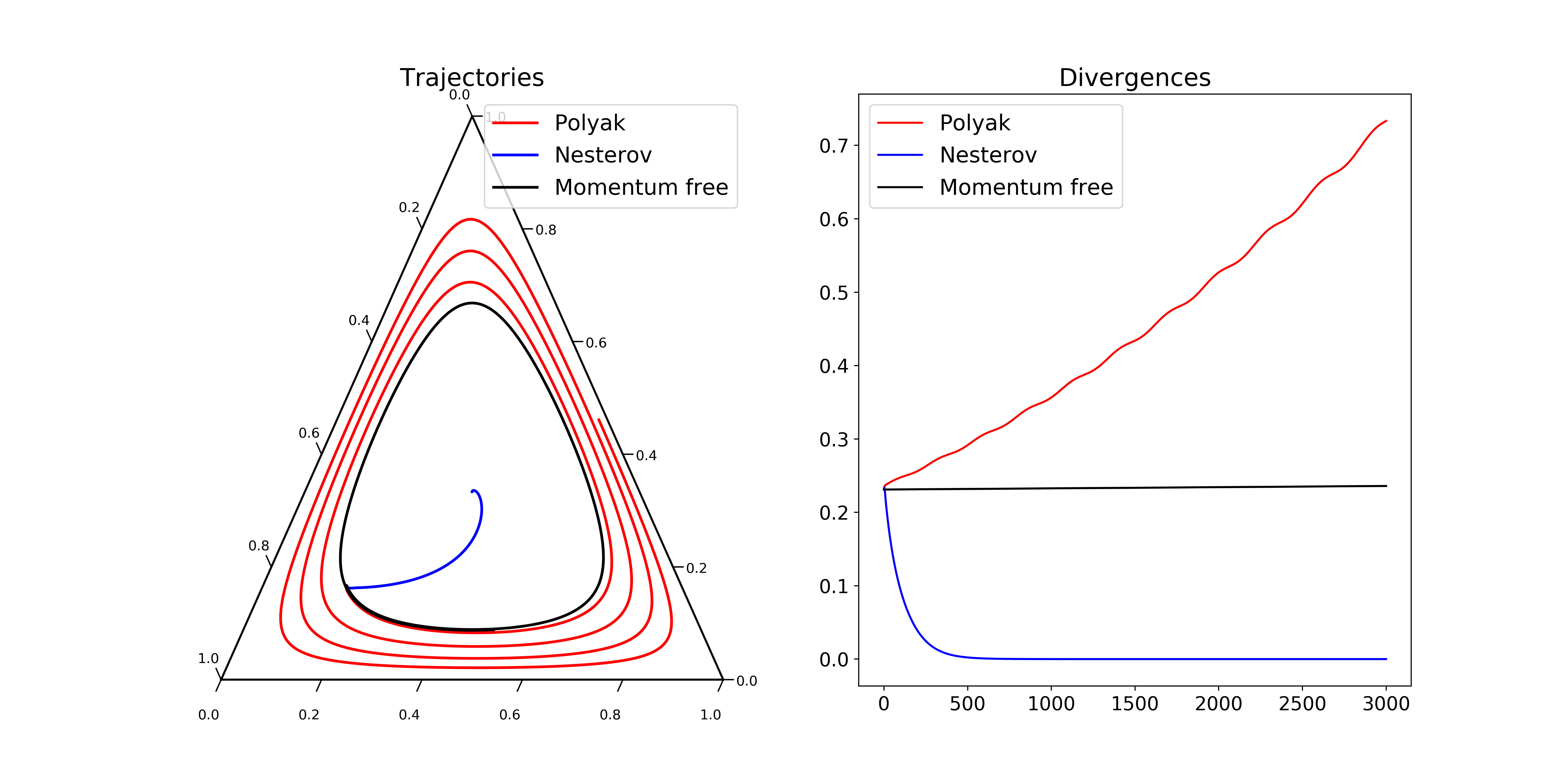}
    \caption{For the rock-paper-scissors landscape ($a=1$, $b=-1$), momentum $\beta=0.65$, and learning rate $\alpha=1/200$, the replicator equation cycles indefinitely with constant KL-divergence based on the initial point. Adding momentum with a non-zero learning rate can cause the cycling to break into either convergence
    or divergence. In this case Nesterov momentum causes the dynamic to converge while Polyak momentum
    causes the dynamic to slowly diverge to the boundary.}
    \label{fig:convergence_divergence}
\end{figure}

\section{Discussion}

We've shown that momentum can accelerate evolutionary dynamics in the probability
simplex just as it does for gradient descent in the machine learning literature.
Lyapunov methods, commonly used to analyze dynamical systems but not yet as commonly
applied in machine learning, allow us to show analytically and explicitly that momentum
decreases the
time to converge for values of momentum typically used in ML, and otherwise
cause divergence or slowdown of trajectories for momentum outside of the interval
$[0, 1)$. Crucially we have shown that learning rate and momentum interact so that
preservation of the convergence properties of the dynamic are guaranteed only for
small $\beta$ or $\alpha$ despite the frequently realized speed up in convergence for larger
values of momentum. 

Interpreting the results, we've shown that the convergence of evolutionary dynamics
can be accelerated by a mechanism of memory that can be viewed as a simple model
of intergenerational information exchange such as epigentics. This may also apply to
immunity or cultural exchanges of information and explain the origin and persistence
of extra-genetic information exchange in lineages and populations.

\subsection{Code}

The code to generate the trajectories and plots in this manuscript is available
as a Python library \textit{pyed} at 
\url{https://github.com/marcharper/pyed}.
Ternary plots were generated with the python-ternary library \cite{pythonternary}.

\begin{ack}
The authors thank Jean Whitmore and Karol Langner for comments on manuscript drafts.

The authors declare no funding sources or competing interests.
\end{ack}

\section*{Broader Impact discussion}

This work provides theoretical insight into the nature of momentum and connections 
between models of evolution and machine learning algorithms. The authors anticipate
no specific immediate ethical implications or societal impacts. This work
may ultimately lead to more efficient machine learning methods and evolutionary
processes, which could have positive or negative consequences depending on the
specific application.

\bibliographystyle{unsrt}
\bibliography{ref}

\newpage

\section{Supplemental Material}

\subsection{Graphical depiction of Theorem \ref{main_theorem}}

\begin{figure}[!htbp]
    \centering
    \includegraphics[width=0.9\textwidth]{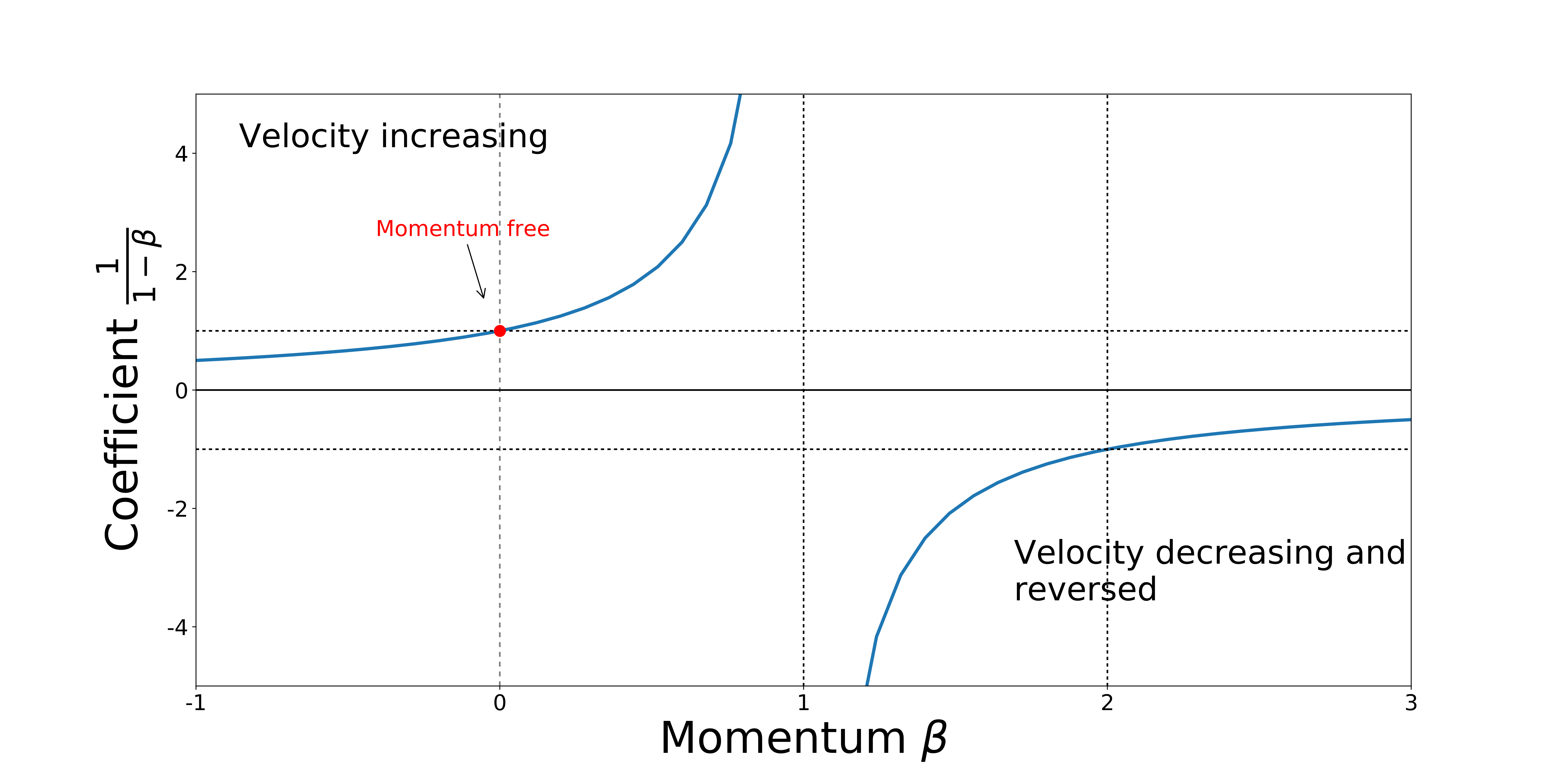}
    \caption{Graphical depiction of Theorem \ref{main_theorem} in terms of the properties of the dynamic coefficient $\frac{1}{1-\beta}$. As $\beta$ varies the convergence and trajectory velocity changes in accordance with the coefficient $1 / (1 - \beta)$. The trajectory velocity is increasing with $\beta$ on $(-\infty, 1)$ and $(1, \infty)$, the
    orientation is reversed on $(1, \infty)$, and the velocity is faster than the momentum free case ($\beta = 0$) for $(0, 1)$ and $(1, 2)$.}
    \label{fig:beta_convergence_impact}
\end{figure}

\subsection{Proof of Equation \ref{heavy_lifter}}

Taking the derivative of the KL-divergence, using the definition of the
continuous replicator equation, gives the following:

\begin{align*}
  \frac{d}{dt}D(\hat{x}||x) &= -\sum_{i}{\frac{\hat{x}_i}{x_i}\frac{dx_i}{dt}}\\
  &= - \sum_i{\hat{x}_i (f_i -\bar{f})} \\
  &= x \cdot f - \hat{x} \cdot f \\
\end{align*}

This quantity is less than zero if $\hat{x}$ is an evolutionarily stable state
(which proves Theorem \ref{lyapunov1} in the continuous case). If we instead
use the replicator equation with momentum (Equation \ref{momentum_replicator}),
a factor of $\frac{1}{1 - \beta}$ is present
on the right hand side of the equation above, proving Equation \ref{heavy_lifter}.

\subsection{Proof of Theorem \ref{momentum_lyapunov}}

\begin{proof}
Since the KL-divergence is positive and zero only at $\ess{x}$, essentially
one just needs to show that the quantity
$$D^{\Delta}(x) = \frac{D(\ess{x}||x') - D(\ess{x}||x)}{\alpha}$$ is less than
zero when $\ess{x}$ is an ESS (for fixed $\alpha$) to establish it as a discrete
Lyapunov function.

A straightforward algebraic calculation shows that this quantity is bounded by
$$ -\log\left( \sum_i \hat{x}_i \frac{x_i'}{x_i} \right)= -\log\left(1 + \alpha \beta \sum_i{\frac{\hat{x}_i z_i} {x_i} } + \alpha \left(\frac{\hat{x} \cdot f}{x \cdot f} - 1\right) \right)$$
which is less than 0 for sufficiently small $\beta$ and the inequality defining an ESS.
\end{proof}

When $\alpha \to 0$ it's easier to directly use ordinary differentiation to prove the continuous
version of Theorem \ref{momentum_lyapunov}.
Proof for the projection dynamic using the Euclidean distance is analogous and omitted.
In the case of Nesterov momentum,
continuity of the fitness landscape and small $\beta$ reduces to the Polyak momentum case.

\end{document}